%% file: main.tex
\pdfoutput=1

\documentclass[letterpaper]{article} 
\usepackage{aaai24}  
\usepackage{times}  
\usepackage{helvet}  
\usepackage{courier}  
\usepackage[hyphens]{url}  
\usepackage{graphicx} 
\usepackage{natbib}  
\usepackage{caption} 
\usepackage{xcolor}
\newcommand\framework{{CMDA}}

\usepackage{newfloat}
\usepackage{listings}
\DeclareCaptionStyle{ruled}{labelfont=normalfont,labelsep=colon,strut=off} 
\usepackage{amsmath}
\usepackage{amssymb}
\usepackage{booktabs}
\usepackage{times}
\usepackage{epsfig}
\usepackage{graphicx}
\usepackage{cuted}
\usepackage{mathtools}
\usepackage{mathabx}
\usepackage{bm}
\usepackage[boxed, ruled,linesnumbered]{algorithm2e}
\usepackage{algpseudocode}
\usepackage{multirow}
\usepackage{makecell}

\title{CMDA: Cross-Modal and Domain Adversarial Adaptation\\for LiDAR-Based 3D Object Detection}

\author{
    Gyusam Chang\textsuperscript{\rm 1}\equalcontrib,
    Wonseok Roh\textsuperscript{\rm 1}\equalcontrib,
    Sujin Jang\textsuperscript{\rm 2},
    Dongwook Lee\textsuperscript{\rm 2},
    Daehyun Ji\textsuperscript{\rm 2},\\
    Gyeongrok Oh\textsuperscript{\rm 1},
    Jinsun Park\textsuperscript{\rm 3},
    Jinkyu Kim\textsuperscript{\rm 4 \dag},
    Sangpil Kim\textsuperscript{\rm 1}\thanks{Corresponding authors.}
}
    
\affiliations{
    \textsuperscript{\rm 1}Department of Artificial Intelligence, Korea University, Republic of Korea\\

    \textsuperscript{\rm 2}Samsung Advanced Institute of Technology (SAIT), Republic of Korea\\

    \textsuperscript{\rm 3}School of Computer Science and Engineering, Pusan National University, Republic of Korea\\

    \textsuperscript{\rm 4}Department of Computer Science and Engineering, Korea University, Republic of Korea\\

    \{gsjang95, paulroh, dhrudfhr98, jinkyukim, spk7\}@korea.ac.kr,\\
    \{s.steve.jang, dw12.lee, derek.ji\}@samsung.com,
    jspark@pusan.ac.kr
}

\def\ie{\emph{i.e.}}
\def\eg{\emph{e.g.}}

\def\etal{\emph{et al.}}
\newcommand{\myparagraph}[1]{\vspace{2pt}\noindent{\bf #1}}

\usepackage{bibentry}

\begin{document}

\maketitle

\input{articles/abstract.tex}
\input{articles/introduction.tex}

\input{articles/related.tex}
\input{articles/methodology.tex}

\input{articles/experiment.tex}
\input{articles/conclusion.tex}

\section{Acknowledgements}
This work was primarily supported by Samsung Advanced Institute of Technology (SAIT) (85\%) and Institute of Information \& communications Technology Planning \& Evaluation (IITP) grant funded by the Korea government (MSIT) (No. 2019-0-00079, Artificial Intelligence Graduate School Program (Korea University), 15\%).

\input{main.bbl}
\end{document}

%% file: articles/abstract.tex

\begin{abstract}
 Recent LiDAR-based 3D Object Detection (3DOD) methods show promising results, but they often do not generalize well to target domains outside the source (or training) data distribution. 
 To reduce such domain gaps and thus to make 3DOD models more generalizable, we introduce a novel unsupervised domain adaptation (UDA) method, called {\framework}, which (i) leverages visual semantic cues from an image modality (i.e., camera images) as an effective semantic bridge to close the domain gap in the cross-modal Bird's Eye View (BEV) representations.
 Further, (ii) we also introduce a self-training-based learning strategy, wherein a model is adversarially trained to generate domain-invariant features, which disrupt the discrimination of whether a feature instance comes from a source or an unseen target domain.
 Overall, our {\framework} framework guides the 3DOD model to generate highly informative and domain-adaptive features for novel data distributions.
 In our extensive experiments with large-scale benchmarks, such as nuScenes, Waymo, and KITTI, those mentioned above provide significant performance gains for UDA tasks, achieving state-of-the-art performance.
\end{abstract}

%% file: articles/introduction.tex
\section{Introduction}
\label{sec:introduction}

3D Object Detection (3DOD) is one of the fundamental computer vision problems and plays a crucial role in real-world applications such as autonomous driving and robotics~\cite{qian20223d, zhu2014single}.
Recent studies~\cite{shi2019pointrcnn, wang2022detr3d, roh2022ora3d} have achieved significant advancements in 3DOD with large-scale benchmarks and precise 3D vision sensors. 
Especially, LiDAR-based approaches~\cite{liang2022bevfusion, liu2023bevfusion, yin2021center} have demonstrated state-of-the-art performance by leveraging precise 3D geometric information (\ie, object location and size) from point clouds.
However, despite these breakthroughs, most 3DOD works face significant performance drops when tested on previously unseen data distributions due to inevitable domain shift issues (\eg, variations in point density, weather conditions, and geographic locations).

To address these challenges, recent approaches in LiDAR-based Unsupervised Domain Adaptation (UDA) primarily focus on effectively leveraging precise geometric information from point clouds~\cite{wang2021pointaugmenting} or self-training strategies with pseudo-labels~\cite{yang2021st3d}.
However, these methods face challenges in learning domain-agnostic contextual information (e.g., colors, textures, and object appearances) relying solely on geometric LiDAR features.

To supplement the absence of semantic information, we introduce Cross-Modality Knowledge Interaction (CMKI), leveraging the contextual details presented in RGB images to guide the learning of rich semantic cues in LiDAR-based geometric features.
Recent studies on multi-modal fusion~\cite{zhang2022cat, bai2022transfusion, liu2023bevfusion} demonstrate that properly complementing 3D point clouds and 2D images with each other enhances overall detection accuracy. 
Due to the individualized multi-modality sensor configuration for each dataset, these methods are still limited in the UDA task. 
To tackle these issues, we advocate for leveraging optimal joint representation, Bird's-Eye-View (BEV), facilitating the transfer of deep semantic clues from 2D image-based features to 3D LiDAR-based features.
Here, the context details of each image pixel can serve as discriminative semantic priors for improved 3DOD performance.
Finally, CMKI enables producing highly informative features by softly associating multi-modal cues. 
We empirically found that our image-assisted approach effectively overcomes domain shift.
To the best of our knowledge, we are the first to adopt the usefulness of multi-modality for UDA on 3DOD.

In addition to utilizing the fine detail of 2D images, we focus on smartly extending the standard self-training approach~\cite{yang2021st3d} to adapt to the previously unseen target data distributions. 
We propose self-training-based learning strategy with Cross-Domain Adversarial Network (CDAN) to relieve the distinct representational gap between source and target data.
To ensure an explicit connection across domains, we first introduce the point cloud mix-up technique, which swaps points sector with random azimuth angles. 
Then, we further apply adversarial regularization to reduce the representational gap across domains, guiding the model to learn domain-invariant information. 
Besides, we design a function that minimizes independent BEV grid-wise entropy to suppress ambiguous and uncertain features derived from mixed inputs. 
Ultimately, our domain-adaptive adversarial self-training approach is now robust in various cross-domain scenarios.

Given landmark datasets in 3DOD, nuScenes~\cite{caesar2020nuscenes}, Waymo~\cite{sun2020scalability}, and KITTI~\cite{Geiger2012CVPR}, we validate the generalizability and effectiveness of our novel UDA framework {\framework}. 
Above all, our proposed framework outperforms the existing state-of-the-art methods on UDA for LiDAR-based 3DOD. 
To summarize, our main contributions are as follows:

\begin{itemize}
\item We propose a novel image-assisted unsupervised domain adaptation approach, called {\framework} with Cross-Modality Knowledge Interaction (CMKI) to yield highly informative features by softly associating multi-modal cues in joint BEV space. To the best of our knowledge, we are the first work that introduces leveraging the semantics of 2D images for UDA on LiDAR-based 3DOD.
\item We design a practical self-training paradigm with Cross-Domain Adversarial Network (CDAN) to relieve the representational gap across domains effectively. Specifically, our approach adversarially constrains the network from learning domain-invariant cues.
\item We analyze the effectiveness of our proposed method on multiple challenging benchmarks, including nuScenes, KITTI and
Waymo. Extensive experiments on various cross-domain adaptation scenarios validate that our proposed method achieves new State-of-the-Art performance for UDA on 3DOD. 
\end{itemize}

\begin{figure*}[t]
    \centering
    \includegraphics[width=\linewidth]{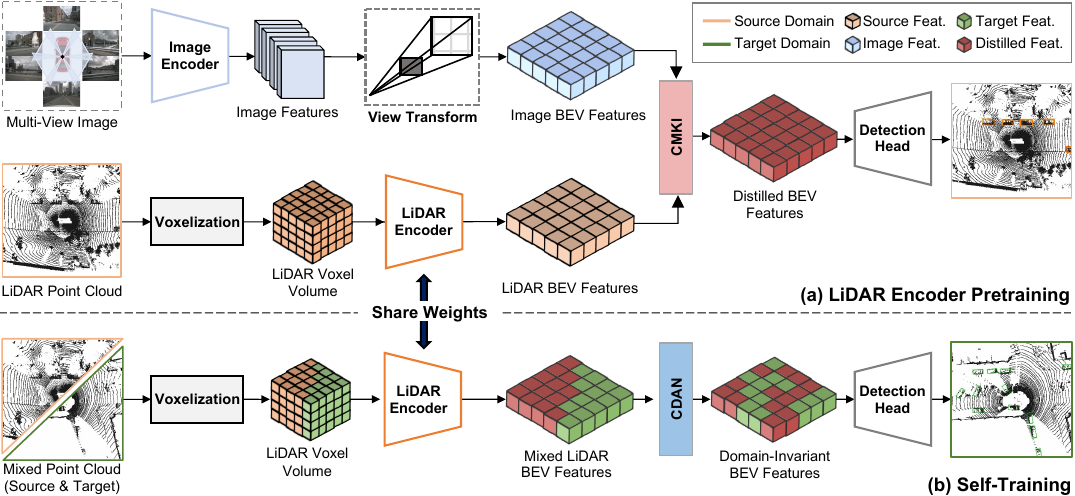}
    \caption{An overview of our architecture. Our framework consists of two main steps.
    (a) Cross-Modal LiDAR Encoder Pre-Training: aligning spatially paired image-based and LiDAR-based BEV representations for cross-modal BEV feature learning.
    This allows the LiDAR encoder to learn modality-specific visual semantic information from the image features.
    (b) Cross-Domain LiDAR-Only Self-Training: learning domain-invariant features through adversarial regularization of the LiDAR encoder, ultimately reducing the representation gap between source and target domains.
    }
    \label{fig:architecture}
\end{figure*}

%% file: articles/related.tex
\section{Related Work}

\label{sec:related}

\subsubsection{LiDAR-based 3D Object Detection.}
In early 3D object detection tasks~\cite{chen2017multi, ku2018joint} which uses point clouds focus on projecting point clouds into 2D feature space by minimizing the loss of spatial information. Recent LiDAR-based 3D object detection works can be categorized as two different approaches: voxel-grid representation and point-based methods. First, the voxel-based approach~\cite{zhou2018voxelnet, yan2018second, yang2018hdnet, lang2019pointpillars, shi2020pv, deng2021voxel} converts point cloud data into voxel representation that is compatible with vanilla Convolution Neural Network(CNN). Also, because the point cloud is sparsely distributed over the whole image, voxel representation constructed from different point sets is more efficient. Although voxel representation is versatile and shows competitive performance in 3D object detection, loss of fine-grained information is inevitable. Differently, to handle this problem, point-based approaches~\cite{yang2019std, shi2019pointrcnn, yang20203dssd} directly use 3D point cloud data to leverage more accurate geometry information than previous methods. In our works, we adopt SECOND~\cite{yan2018second} and PV-RCNN~\cite{shi2020pv} as baseline models that are the representative networks in 3D object detection to demonstrate the effectiveness of extracting domain invariant features with our proposed methods. 

\subsubsection{Unsupervised Domain Adaptation (UDA) for LiDAR-based 3D Object Detection.}
To generalize LiDAR-based 3D Object Detection for autonomous driving, Unsupervised Domain Adaptation addresses performance drop between a labeled source dataset and an unlabeled target dataset. In early UDA for LiDAR-based 3D object detection, Y. Wang \etal~\cite{wang2020train} propose to mitigate the inductive bias of box scale by unfamiliar objects exploiting Statistical Normalization (SN). ST3D~\cite{yang2021st3d} applied Random Object Scaling (ROS) and a novel self-training framework in the data pipeline to demonstrate efficiency in the target scenario. Turning to the domain of point cloud resolution, recent studies suggest various methods to complement the sparsity of point clouds. SPG~\cite{xu2021spg} enriches the missing points by employing efficient point generation. 3D-CoCo~\cite{yihan2021learning} utilizes domain alignment between source and target to extract robust features from unlabeled point clouds. LiDAR Distillation~\cite{wei2022lidar} generates pseudo sparse point sets leveraging spherical coordinates and transfers the knowledge of the source, effectively reducing the domain gap. 

%% file: articles/methodology.tex
\section{Method}
\label{sec:methodology}

In this section, we present a novel Unsupervised Domain Adaptation (UDA) framework CMDA for LiDAR-based 3D object detection (3DOD).
We advocate leveraging multi-modal inputs during the training phase to enhance the generalizability across diverse domains.
Specifically, we encourage the LiDAR BEV features to learn rich-semantic knowledge from camera BEV features and explicitly guide such cross-modal learning via cross-domain adversarial pipeline, achieving generalized perception against unseen target conditions. 
We first provide an overview of our framework and then present technical details of the proposed methods as follows: (1) Cross-Modality Knowledge Interaction (CMKI) and (2) Cross-Domain Adversarial Network (CDAN).

\subsection{Overview}\label{sec:3-1}
We illustrate an overview of our framework in Alg.~\ref{alg:overview} and Fig.~\ref{fig:architecture}, which aims to maximize the potential of each modality in guiding the 3DOD models for improved generalizability.
During the training phase, the model takes multi-view images $I = \{i_1, i_2, ... , i_{N^{I}}\} \in \mathbb{R}^{{N^{I}} \times H \times W \times 3}$ and 3D point clouds $P = \{p_1, p_2, ... , p_{N^{P}}\} \in \mathbb{R}^{{N^{P}} \times 3}$ as inputs, and outputs a set of 3D bounding boxes $\hat{L}$.
Our primary goal is to effectively transfer a LiDAR-based 3DOD model trained on labeled source domain data ${\{(P_i^s, I_i^s, L_i^s)\}}_{i=1}^{N_s}$ to the unlabeled target domain data ${\{(P_i^t, L_i^t)\}}_{i=1}^{N_t}$.
Here, $P_i^s, I_i^s$, and $L_i^s$ represent the $i$-th point clouds, multi-view images, and their corresponding ground truth labels from the source domain.
Similarly, $P_i^t$ and $L_i^t$ denote the $i$-th point clouds and their corresponding pseudo-label from the target domain.
$N_s$ and $N_t$ indicate the number of samples from the source and the target domain, respectively. 
Note that we do not use the target domain labels during training.

\subsection{Cross-Modality Knowledge Interaction (CMKI)}\label{sec:3-2}
Recently, multi-modal 3DOD mechanisms~\cite{zhang2022cat, bai2022transfusion, vora2020pointpainting} have highlighted the benefits of synergistically complementing geometric losses from 2D images with semantic losses from 3D point clouds.
Despite their potential benefits, the introduction of multi-modalities to address both geometric and semantic domain gaps has received limited attention in the field of UDA for 3DOD tasks.
In this work, we delve into the effectiveness of interactions between different modalities (\ie, camera images and LiDAR point clouds), aiming to enhance the BEV feature quality and improve detection performance on previously unseen target data distributions.

\subsubsection{Optimal Joint Representation.}
Precise geometric alignment is essential to ensure the quality of both image and point cloud features.
Although existing association techniques~\cite{focalsconv-chen, vora2020pointpainting} with calibration matrices employ multi-modal information, they do not fully take advantage of the deep semantic clues from the images due to the non-homogeneous nature of features and their representations.
For example, camera features are encoded in single or multiple-perspective views, whereas LiDAR features are expressed in the BEV space~\cite{bai2022transfusion}.
Hence, we are motivated to find an optimal joint representation to facilitate effective cross-modal knowledge interaction and to investigate its impact on the UDA task. 
Inspired by recent multi-modal fusion~\cite{liu2023bevfusion, liang2022bevfusion}, we adopt BEV feature representations and aim to transfer valuable modality-specific cues between them.

\SetKwComment{Comment}{\# }{}
\RestyleAlgo{ruled}
\setlength{\textfloatsep}{8pt}
\begin{algorithm}[tb]
	\setlength{\belowcaptionskip}{-20pt}
	\caption{Overview of our framework {\framework}.}
	\label{alg:overview}  
        \Indm 
        \KwIn{
            Source labeled data $\{(P^s_i, I^s_i, L^s_i)\}^{N_s}_{i=1}$ and target pseudo-labeled data $\{(P_i^t, L_i^t)\}_{i=1}^{N_t}$.
        }
	\KwResult{
            Robust 3D detector for the target domain.
            
            \textbf{Procedure:}
        }
        
        \Comment*[h]{LiDAR Encoder Pretraining.}
        
        \Indp
        \While{$i = N_s$}{
        
        Transform 3D points $P^s_i$ to BEV feature $F_{P}^{bev}$.

        Transform 2D images $I^s_i$ to BEV feature $F_{I}^{bev}$.
        
        Guide $F_{P}^{bev}$ to contain semantic clues from $F_{I}^{bev}$.
 
        }
        \Indm\Comment*[h]{Self-Training.}
        
        \Indp
        \While{$i = N_t$}{ 
        Generate pseudo label $L^t_i$ for self-training.
        
        Mix source point sector with target point sector according to Eq.~\ref{eq:mixup}.
        
        Extract instance-level features ${\bf f}_i$ using mixed point $P^{mix}$.
        
        Generate Domain labels $y_{r}$ based on location.

        Close the representational gap using adversarial discriminator $\phi_D$.
        }
        
\end{algorithm}


\subsubsection{Cross-Modal BEV Feature Map Generation.} 
\begin{figure}[t]
  \centering
  \includegraphics[width=\linewidth]{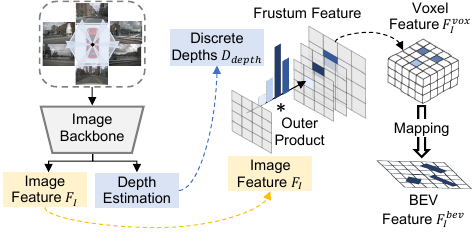}
  \caption{An overview of our Images-to-BEV View Transform module. We first transform multi-view images into voxel-wise representations $F_{I}^{vox}$ by simultaneously leveraging $F_I$ and $D_{depth}$, yielding a BEV representation $F_{I}^{bev}$.}
  \label{fig:imgtobev}
\end{figure}

Inspired by the prevalent work, Lift-Splat-Shoot (LSS)~\cite{philion2020lift}, our camera stream (illustrated in Fig.~\ref{fig:imgtobev}) transforms RGB images into high-level BEV representations.
First, the image encoder extracts rich-semantic visual features $F_{I} \in \mathbb{R}^{{N^{I}} \times H \times W \times C}$ from the multi-view images $I \in \mathbb{R}^{{N^{I}} \times H \times W \times 3}$. 
To construct the BEV feature, we apply a view transform module that links the 2D image coordinate to the 3D world coordinate. 
For each pixel, we densely predict representations at all possible depths $D_{depth} \in \mathbb{R}^{H \times W \times D}$ in a classification manner, where $D$ denotes the discrete depth bins.
We then complete the frustum-shaped voxels of contextual features by calculating the outer product of $D_{depth}$ and $F_{I}$.
Given the camera parameters, we obtain a pseudo voxel via the interpolation process, which is fed into a voxel backbone to extract features $F_{I}^{vox} \in \mathbb{R}^{X \times Y \times Z \times C}$.
Then, $F_{I}^{vox}$ are compressed along the height axis to yield the image-based BEV feature map $F_{I}^{bev} \in \mathbb{R}^{X \times Y \times ZC}$.

To generate the BEV feature map $F_{P}^{bev} \in \mathbb{R}^{X \times Y \times ZC}$ from the 3D LiDAR point clouds, we follow standard voxel-based height compression method~\cite{zhou2018voxelnet}.

\subsubsection{Cross-Modal Knowledge Interaction in BEV Features.} 
As reported by previous studies~\cite{wang2020train, yang2021st3d, xu2021spg, wei2022lidar}, UDA performance is significantly enhanced by leveraging precise geometric details from 3D point cloud data to guide feature-level adaptation.
Although point clouds provide geometrically informative cues, they are limited in generating rich semantic information such as colors, textures, and the appearance of target objects and backgrounds.
To complement such a lack of contextual information, we are motivated to exploit the fine detail of RGB images as discriminative semantic priors for improved 3DOD performance.
To this end, we focus on transferring the rich semantic knowledge from image-based features to LiDAR-based features.
Based on the joint BEV representations between modalities, we formulate cross-modal knowledge interaction with:
\begin{equation}
	\mathcal{L}_{cmki} = \frac{1}{XY} \sum_{i=1}^X\sum_{j=1}^Y \left\lVert F_{P}^{bev}(i,j) - F_{I}^{bev}(i,j) \right\rVert_{2},
\end{equation}
where $X, Y$ denotes the width and length of the BEV feature map and $\left\lVert\cdot\right\rVert_{2}$ is the $L_2$ norm. 
By minimizing $\mathcal{L}_{cmki}$, we optimize 3D LiDAR-based features to contain highly informative semantic clues from 2D image-based features.
Our BEV-based cross-modal knowledge interaction establishes valuable connections between input modalities and consistently yields improvements across various cross-domain deployments, as demonstrated in Tables~\ref{tab:SOTAcomparison} and ~\ref{tab:new_ablation}.

\subsection{Cross-Domain Adversarial Network (CDAN)}\label{sec:3-3}

\begin{figure}[t]
  \centering
  \includegraphics[width=\linewidth]{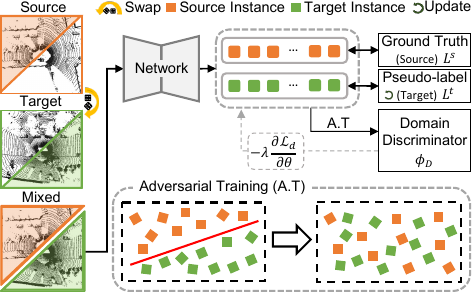}
  \caption{An overview of our cross-domain self-training step. Given a mixed point scene (source-domain points replace target-domain points in a randomly chosen region), our domain discriminator is adversarially trained to classify whether an object is from source or target domains.}
  \label{fig:selftraining}
\end{figure}

In the field of UDA, self-training strategies~\cite{xie2020self, yang2021st3d, yihan2021learning, zou2018unsupervised} with target pseudo-labels have significantly enhanced the performance of 3DOD models on unsupervised environments.
However, we empirically discover that addressing the representational gap between source and target domains still poses challenges.
Concretely, these approaches struggle to accurately recognize target objects composed of less familiar point samples and lead to generating low-quality target pseudo-labels as illustrated in Fig.~\ref{fig:graph}(a).
To this end, we propose a domain-adaptive adversarial self-training approach, as shown in Fig.~\ref{fig:selftraining}, to enhance the learning of domain-agnostic features and improve the accuracy of pseudo-labels. 

\subsubsection{Cross-Domain Mix-up.}
To reduce the distributional shift between domains during the self-training process, we first switch the point cloud sectors of the source and target domain scenes.
We cut both point cloud sectors and corresponding labels at the identical azimuth angle and swap them with each other, following~\cite{xiao2022polarmix}.
Note that the azimuth angle $\mathcal{\theta}$ is randomly set within a specific range for each iteration to avoid inductive biases. 
Our mix-up process is formulated as follows:
\begin{equation}
\begin{aligned}
    (P^{mix}, L^{mix}) &= M_{s}^{\theta}(P^s, L^s) \oplus M_{t}^{2\pi-\theta}(P^t, L^t),
\end{aligned}
\label{eq:mixup}
\end{equation}
where $M_{s}$ and $M_{t}$ denote binary mask to filter points within the azimuth angle $\mathcal{\theta}$; and $L^t$ represents target pseudo-labels. 
Next, the pre-trained encoder takes the mixed point cloud $P^{mix}$ as input and generates LiDAR-based BEV features.
Our mix-up strategy directly introduces cross-domain data instances to facilitate learning domain-invariant features, ultimately leading to improved adaptation performance.

\subsubsection{Domain Adaptive Discriminator.}
We introduce the domain adaptive adversarial discriminator to implicitly reduce the representational gap between the source and the target within the shared embedding space.
Specifically, we guide the detection head to learn generalized information during the self-training through an adversarial learning paradigm using Gradient Reversal Layer~\cite{ganin2016domain}.
Our cross-domain discriminator $\phi_D$ tries to classify the domain of instance-level features ${\bf f}_i$ for $i=\{1, 2, \dots, |N_f|\}$ from the detection head.
Each feature instance is labeled with $y_r$ based on its coordinates to indicate whether it belongs to the source or target regions.
We train $\phi_D$ to discriminate the domain of each instance based on the following loss function:
\begin{equation}
{\mathcal{L}}_{d} = -\mathbb{E}_{f, y_r\sim\mathbb{D}}\Bigg[\sum_{r\in\mathcal{R}}y_{r} \log \phi_D(f)_{r}\Bigg],
\end{equation}
where $\mathbb{E}_{f, y_r\sim\mathbb{D}}$ indicates an expectation over samples $(f, y_r)$ drawn from the input data distribution $\mathbb{D}$. 
While the discriminator $\phi_D$ is trained to identify the domain of each instance accurately, the 3DOD model produces instance-level features to fool the discriminator in distinguishing their domains (\ie, negative loss function $-\mathcal{L}_d$).
This way, we use adversarial guidance to encourage the 3DOD model to learn domain-agnostic features.
Furthermore, in order to mitigate ambiguous features induced by randomly mixed domain scenes and enhance prediction confidence, we regularize the network using BEV grid-wise entropy loss $\mathcal{L}_{ent}$:
\begin{equation}
\label{loss:ent}
{\mathcal{L}}_{ent} = \frac{-1}{\log{ZC}}\sum_{i=1}^{X}\sum_{j=1}^{Y}\sum_{c=1}^{ZC}F_{P}^{bev}(i,j,c)\log{F_{P}^{bev}(i,j,c)}
\end{equation}
Finally, we advance a simple self-training stage to improve generalizability across various target domains with the following loss $\mathcal{L}_{cdan}$ as a sum of the two losses $\mathcal{L}_{d}$ and $\mathcal{L}_{ent}$:
\begin{equation}
    \mathcal{L}_{cdan} = \lambda_{d}\mathcal{L}_{d} + \lambda_{ent}\mathcal{L}_{ent}
    \label{eq:loss-da}
\end{equation}

\subsubsection{Loss Function.} 
We also leverage conventional loss term $\mathcal{L}_{det}$ associated with regression of 3D bounding box parameters and classification of object categories.
Taking all loss functions together, our learning objective is:
\begin{equation}
    \mathcal{L}_{total} = \lambda_{det}\mathcal{L}_{det} + \mathcal{T}_{cmki}\lambda_{cmki}\mathcal{L}_{cmki} + \mathcal{T}_{cdan}\lambda_{cdan}\mathcal{L}_{cdan}
\end{equation}
where $\lambda$ is a hyperparameter derived from grid searches to handle the strength of each loss term. In addition, we employ the binary toggle $\mathcal{T}$ for 
carefully scheduled training processes, where $(\mathcal{T}_{cmki}, \mathcal{T}_{cdan})$ set $(1, 0)$ for source training and $(0, 1)$ for self-training.

%% file: articles/experiment.tex

\renewcommand{\arraystretch}{1.3} 
\begin{table*}[t]
	\begin{center}
    	\resizebox{\linewidth}{!}{%
    	\begin{tabular}{@{}clcccc@{}} \toprule
            \multirow{2.3}{*}{Task} & \multirow{2.3}{*}{Model}  & \multicolumn{2}{c}{SECOND-IoU~\cite{yan2018second}} & \multicolumn{2}{c}{PV-RCNN~\cite{shi2020pv}} \\\cmidrule{3-6}
            &  & $\text{BEV~AP}$ $\uparrow$ / $\text{3D~AP}$ $\uparrow$ & Closed Gap $\uparrow$ & $\text{BEV~AP}$ $\uparrow$ / $\text{3D~AP}$ $\uparrow$ & Closed Gap $\uparrow$ \\
            \midrule
            \multirow{5}{*}{\makecell{nuScenes \\ $\rightarrow$ Waymo}} & Direct Transfer & 39.18 / 20.78  &  & 41.30 / 25.89 &  \\
            & ST3D~\cite{yang2021st3d}  & 45.35 / 27.12 & +21.62\% / +19.08\% & 52.50 / 36.21 & +38.63\% / +31.07\% \\
            & ST3D++~\cite{yang2022st3d++}  & 44.87 / 25.79 & +19.94\% / +15.08\% & --.-- / --.-- & --.--\% / --.--\% \\
            & {\framework} (Ours)  & \textbf{46.79} / \textbf{29.42} & \textbf{+26.66}\% / \textbf{+26.00}\% & \textbf{58.57} / \textbf{45.58} & \textbf{+59.57}\% / \textbf{+59.29}\% \\\cmidrule{2-6}
            & Oracle & 67.72 / 54.01 &  & 70.29 / 59.10 &  \\
            \midrule
            \multirow{7}{*}{\makecell{nuScenes \\ $\rightarrow$ KITTI}} 
            & Direct Transfer & 51.84 / 17.92   &  & 68.15 / 37.17 &  \\
            & SN~\cite{wang2020train} & 40.03 / 21.23 & {\large-}37.55\% / +05.96\%  & 60.48 / 49.47 & {\large-}36.82\% / +27.13\% \\
            & ST3D~\cite{yang2021st3d}  & 75.94 / 54.13 & +76.63\% / +59.50\% & 78.36 / 70.85 & +49.02\% / +74.30\% \\
            & ST3D++~\cite{yang2022st3d++}  & 80.52 / 62.37 & +91.19\% / +80.05\% & --.-- / --.-- & --.--\% / --.--\% \\
            & DTS~\cite{hu2023density} & 81.40 / 66.60 & +93.99\% / +87.66\% & 83.90 / 71.80 & +75.61\% / +76.40\% \\
            & {\framework} (Ours) & \textbf{82.13} / \textbf{68.95} & \textbf{+96.31}\% / \textbf{+91.90}\% & \textbf{84.85} / \textbf{75.02} & \textbf{+80.17}\% / \textbf{+83.50}\% \\\cmidrule{2-6}
            & Oracle & 83.29 / 73.45 &  & 88.98 / 82.50 &  \\
            \midrule
            \multirow{8}{*}{\makecell{Waymo \\ $\rightarrow$ nuScenes}} 
            & Direct Transfer & 32.91 / 17.24  &  & 34.50 / 21.47 &  \\
            & SN~\cite{wang2020train} & 33.23 / 18.57 & +01.69\% / +07.54\%  & 34.22 / 22.29 & {\large-}01.50\% / +04.80\% \\
            & ST3D~\cite{yang2021st3d}  & 35.92 / 20.19 & +15.87\% / +16.73\% & 36.42 / 22.99 & +10.32\% / +08.89\% \\
            & ST3D++~\cite{yang2022st3d++}  & 35.73 / 20.90 & +14.87\% / +20.76\% & --.-- / --.-- & --.--\% / --.--\% \\
            & LD~\cite{wei2022lidar}  & 40.66 / 22.86 & +40.85\% / +31.88\% & 43.31 / 25.63 & +47.34\% / +24.34\% \\
            & DTS~\cite{hu2023density}  & 41.20 / 23.00 & +43.70\% / +32.67\% & 44.00 / 26.20 & +51.04\% / +27.68\% \\
            & {\framework} (Ours w/ LD) & \textbf{42.81} / \textbf{24.64} & \textbf{+52.19}\% / \textbf{+41.97}\% & \textbf{44.44} / \textbf{26.41} & \textbf{+53.41}\% / \textbf{+28.91}\% \\\cmidrule{2-6}
            & Oracle & 51.88 / 34.87 &  & 53.11 / 38.56 &  \\
           \bottomrule
        \end{tabular}}
        \caption{Comparisons of Unsupervised Domain Adaptation (UDA) performance with state-of-the-art approaches, including SN~\cite{wang2020train}, ST3D~\cite{yang2021st3d}, ST3D++~\cite{yang2022st3d++}, LiDAR Distillation (LD)~\cite{wei2022lidar} and DTS~\cite{hu2023density}. 
        For fair comparisons, we train our LiDAR-based object detector with two baseline methods: SECOND~\cite{yan2018second} and PV-RCNN~\cite{shi2020pv}.
        We report UDA performance in three popular benchmarks: nuScenes~\cite{caesar2020nuscenes} $\rightarrow$ Waymo~\cite{sun2020scalability}, nuScenes $\rightarrow$ KITTI~\cite{Geiger2012CVPR}, and Waymo $\rightarrow$ nuScenes. Evaluation metrics include moderate $\text{BEV~AP}$ and $\text{3D~AP}$ (IoU threshold=0.7) and Closed Gap for car objects. 
        }
        \label{tab:SOTAcomparison}
    \end{center}
\end{table*}

\section{Experiments}
\label{sec:experiment}

\subsubsection{Datasets.}
We evaluate overall performance on landmark datasets for 3D object detection task: nuScenes~\cite{caesar2020nuscenes}, Waymo~\cite{sun2020scalability}, and KITTI~\cite{Geiger2012CVPR}. 
The three datasets have different point cloud ranges and specifications. Hence, we convert them to a unified range $[-75.2, -75.2, -2, 75.2, 75.2, 4]$ and adopt only seven parameters to achieve consistent training results under the same conditions: center locations $(x, y, z)$, box size $(l, w, h)$, and heading angle $\delta$. 

\subsubsection{Evaluation Metrics.}
We follow the KITTI evaluation metric for consistent evaluation across datasets.
Also, we adopt the 360-degree surrounding view configuration for evaluation, apart from the KITTI dataset, which only offers the annotations in the front view.
We report the Average Precision (AP) over 40 recall positions and 0.7 IoU thresholds for both the BEV IoUs and 3D IoUs.
To offer empirical lower and upper bounds on adaptation performance, we present three additional reference points: $\textbf{Direct Transfer}$—evaluating the source domain pre-trained model directly on the target domain, $\textbf{Oracle}$—the fully supervised model trained on the target domain, and $\textbf{Closed Gap}$—representing the hypothetical closed gap by
\begin{equation}
    \label{metric:closed_gap}
    \text{Closed Gap}=\frac{\text{AP}_{\text{model}} - \text{AP}_{\text{Direct Transfer}}}{\text{AP}_{\text{Oracle}} - \text{AP}_{\text{Direct Transfer}}} \times 100\%.
\end{equation}


\myparagraph{Performance Comparison with SOTA Approaches.}\label{exp:4-2}
As shown in Tab.~\ref{tab:SOTAcomparison}, we quantitatively compare our proposed framework with existing state-of-the-art methods, which include Statistical Normalization (SN)~\cite{wang2020train}, ST3D~\cite{yang2021st3d}, ST3D++~\cite{yang2022st3d++}, LiDAR Distillation (LD)~\cite{wei2022lidar}, and DTS~\cite{hu2023density}.
SN applies statistical normalization to reduce the inductive bias of box scales in the cross-domain setting. ST3D improves the effectiveness of the self-training process with data augmentation, while LD mitigates the beam-induced dense-to-sparse density shift by generating pseudo points. These methods demonstrate notable capacity but still rely on geometric information from 3D sensors and often face challenges in effectively adapting to unseen target domains. 
To overcome these limitations, we introduce \framework, featuring Cross-Modality Knowledge Interaction (CMKI) and Cross-Domain Adversarial Network (CDAN).

In Tab.~\ref{tab:SOTAcomparison}, we observe that our {\framework} generally outperforms the other five methods in all metrics, including BEV AP, 3D AP, and Closed Gap.
Following existing work, we evaluate UDA performance in three different scenarios: (1) nuScenes~\cite{caesar2020nuscenes} $\rightarrow$ Waymo~\cite{sun2020scalability}, (2) nuScenes $\rightarrow$ KITTI~\cite{Geiger2012CVPR}, and (3) Waymo $\rightarrow$ nuScenes. 
The performance gain of {\framework} is more apparent in scenarios (1) and (3), which utilize multi-view camera images, and thus benefit from highly instructive visual details. 
More importantly, in the dense to sparse subdomain shift setting, \ie, Waymo $\rightarrow$ nuScenes, {\framework} achieves a substantial performance improvement of \textit{Closed Gap},
by up to  +52.19\%/+41.97\% on SECOND-IoU~\cite{yan2018second}, and +53.41\%/+28.91\% on PV-RCNN~\cite{shi2020pv} for BEV AP / 3D AP. These promising scores demonstrate that our framework can effectively boost 3DOD performance in the unsupervised target domain, even with fewer LiDAR sensor beams.
Note that SN and LD are unsuitable for nuScenes $\rightarrow$ Waymo task and are therefore excluded for a fair comparison.

Remarkably, our {\framework} framework also achieves higher adaptation scores when utilizing single-view camera images, \ie, nuScenes $\rightarrow$ KITTI.
In this case, {\framework} with SECOND-IoU achieves +96.31\% / +91.90\% of \textit{Closed Gap}. 
Overall, our {\framework} framework effectively reduces the distributional shift between the source and target domains, leading to new state-of-the-art performance.

\subsection{Qualitative Analyses}\label{exp:4-5}

\begin{figure}[ht]
    \centering
    \includegraphics[width=\linewidth]{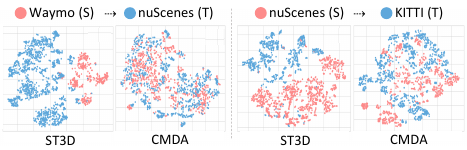}
    \caption{t-SNE~\cite{van2008visualizing} visualizations of source (S, red) and target (T, blue) domains' LiDAR-based BEV feature distribution. 
    }
    \label{fig:tsne}
\end{figure}

\myparagraph{t-SNE Analysis.}
To assess the extent of the domain gap, we provide t-SNE~\cite{van2008visualizing} visualizations of the feature space learned from both the source (red) and target (blue) domains.
As shown in Fig.~\ref{fig:tsne}, ST3D exhibits distinct clusters for the source and target domains, whereas CMDA results in a harmoniously dispersed feature space encompassing both target and source domains. 
These qualitative findings confirm that CMDA effectively encourages the model to learn domain-invariant features.

\myparagraph{Impact of Utilizing Visual Semantic Priors.}
\begin{figure}
    \centering
    \includegraphics[width=\linewidth]{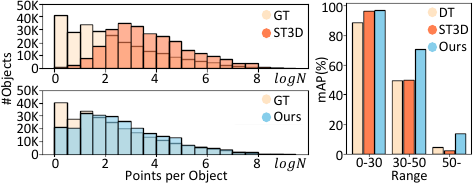}
    \caption{Statistical analyses of detection results: (left) perception capacity for the number of points per object and (right) accuracy comparison across the range.
    }
    \label{fig:graph}
\end{figure}
\begin{figure}
    \centering
    \includegraphics[width=\linewidth]{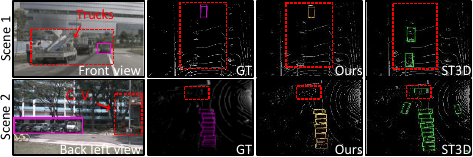}
    \caption{Qualitative visualization of  Waymo $\rightarrow$ nuScenes adaptation. Magenta, green, and yellow represent Ground Truth, ST3D, and Ours. For better understanding, we visualize corresponding camera views along with the red dotted line showing the region where the domain shift is prominent.
    }
    \label{fig:visualization}
\end{figure}
To validate the effectiveness of the semantic priors learned from the image-based BEV features, we present additional experimental results and qualitative analyses.
In Fig.~\ref{fig:graph} (left), we perform a statistical evaluation of the perception capacity based on various point densities per object.
{\framework} effectively detects objects even with relatively sparse points. 
Fig.~\ref{fig:graph} (right) shows that {\framework} achieves improved detection accuracy (mAP\%), particularly for distant objects.

Further, Scene 1 of Fig.~\ref{fig:visualization} provides a notable example of improved detection accuracy for distant objects. ST3D~\cite{yang2021st3d} fails to detect a relatively distant object, whereas {\framework} successfully detects it.
Also, in Scenes 1 and 2, ST3D struggles to adapt from uniform-labeled (vehicle) to various-labeled (car, truck, bus, construction vehicle, etc.) domains.
In contrast, ours effectively discriminates ``cars'' from ``construction vehicles'' and ``trucks''. 
These findings confirm the effectiveness of utilizing the visual semantic priors jointly learned from image-based features to improve the overall UDA performance. 

\subsection{Ablation Studies}\label{exp:4-3}
\begin{table}[t]
	\begin{center}
        \resizebox{\linewidth}{!}{%
        \begin{tabular}{@{}cccccc@{}} \toprule
            \multirow{2.5}{*}{Task} & \multicolumn{3}{c}{Method} & SECOND-IoU & PV-RCNN  \\\cmidrule{5-6}
            & ST3D & CMKI & CDAN & BEV AP / 3D AP & BEV AP / 3D AP \\\midrule
            \multirow{5}{*}{\makecell{nuScenes \\ $\rightarrow$ Waymo}}
            & - & - & - & 39.18 / 20.78 & 41.30 / 25.89\\
            & - & $\checkmark$ & - & 44.41 / 22.26 & 47.28 / 28.57 \\
            & $\checkmark$ & - & - & 45.35 / 27.12 & 52.50 / 36.21\\
            & $\checkmark$ & $\checkmark$ & - & 45.43 / 28.63 & 53.04 / 42.75\\
            & $\checkmark$ & $\checkmark$ & $\checkmark$ & \textbf{46.79} / \textbf{29.42} & \textbf{58.57} / \textbf{45.58}\\\midrule
            \multirow{5}{*}{\makecell{nuScenes \\ $\rightarrow$ KITTI}}
            & - & - & - & 51.84 / 17.92 & 68.15 / 37.17\\
            & - & $\checkmark$ & - & 63.86 / 37.22 & 72.12 / 40.17\\
            & $\checkmark$ & - & - & 75.94 / 54.13 & 78.36 / 70.85\\
            & $\checkmark$ & $\checkmark$ & - & 78.52 / 60.04 & 82.43 / 72.20\\
            & $\checkmark$ & $\checkmark$ & $\checkmark$ & \textbf{82.13} / \textbf{68.95} & \textbf{84.85} / \textbf{75.02} \\
            \bottomrule
        \end{tabular}}
        \caption{Ablation study to see the effect of CMKI and CDAN. We report BEV AP and 3D AP (IoU=0.7) in the following domain adaption scenarios: (i) nuScenes~\cite{caesar2020nuscenes} $\rightarrow$ Waymo~\cite{sun2020scalability} and (ii) nuScenes $\rightarrow$ KITTI~\cite{Geiger2012CVPR}.
        }
        \label{tab:new_ablation}
     \end{center}
\end{table}

            
%

\myparagraph{Effect of CDAN and CMKI.}
In Tab.~\ref{tab:new_ablation}, we evaluate CMKI and CDAN in various adaptation configurations on SECOND-IoU~\cite{yan2018second} and PV-RCNN~\cite{shi2020pv}.
ST3D~\cite{yang2021st3d} denotes a baseline self-training method, and the first row in each setting indicates \textit{Direct Transfer}.
To investigate the sole effect of each approach, we deliberately did not use any augmentation strategies (\eg, LD~\cite{wei2022lidar}, SN~\cite{wang2020train}).
Tab.~\ref{tab:new_ablation} demonstrates that the addition of CMKI and CDAN improves adaptive capability across all experiments. 
Notably, CMKI emphasizes the significance of rich semantic knowledge in achieving generalized recognition, narrowing the gaps by up to +12.02\% in BEV AP and +19.30\% in 3D AP, when compared to \textit{Direct Transfer}.
CDAN further enhances the generalizability by learning domain-agnostic BEV features through the adversarial discriminator, achieving improvements of up to +30.29\% in BEV AP and +51.03\% in 3D AP compared to \textit{Direct Transfer}.
These results prove the power of our {\framework} framework in substantially enhancing the quality of UDA for 3DOD.

\renewcommand{\arraystretch}{1.3} 
\begin{table}[t]
	\begin{center}
        \resizebox{\linewidth}{!}{%
    	\begin{tabular}{@{}cccccc@{}} \toprule
            \multirow{2}{*}{Task} & \multicolumn{3}{c}{Method} & {SECOND-IoU} & {PV-RCNN}  \\\cmidrule{5-6}
            & ~ST3D & CL & CDAN & BEV AP $\uparrow$ / 3D AP $\uparrow$ & BEV AP $\uparrow$ / 3D AP $\uparrow$ \\\midrule
            \multirow{4}{*}{\makecell{nuScenes \\ $\rightarrow$ KITTI}}~ & - & - & - & 51.84 / 17.92 & 68.15 / 37.17 \\
            & $\checkmark$ & - & - & 75.94 / 54.13 & 78.36 / 70.85 \\
            & $\checkmark$ & $\checkmark$ & - & 79.20 / 58.20 & 80.99 / 71.51 \\
            & $\checkmark$ & - & $\checkmark$ & \textbf{80.13} / \textbf{63.67} & \textbf{83.27} / \textbf{73.05} \\
           \bottomrule
        \end{tabular}}
        \caption{Comparisons of domain adaptation performance with Contrastive Learning (CL) approach in nuScenes~\cite{caesar2020nuscenes} $\rightarrow$ KITTI~\cite{Geiger2012CVPR}.}
        \label{tab:DAvsCL}
     \end{center}
\end{table}


\myparagraph{CDAN vs. Contrastive Learning (CL).}
To validate the effectiveness of CDAN, in Tab.~\ref{tab:DAvsCL}, we provide a comparison with Contrastive Learning (CL)-based adaptation approach following 3D-CoCo~\cite{yihan2021learning}. 
For a fair comparison, we employ the identical source pre-trained weights and apply each learning strategy on instances from the detection head during self-training.
While the CL-based method, with well-matched positive/negative pairs, enhances the baseline self-training approach (ST3D), it exhibits limited improvements compared to CDAN due to implicit issues such as sample discrepancy or precision errors.
Unlike the CL-based approach, CDAN benefits significantly from adversarial mechanism and successfully tackle these challenges, producing stable adaptation effects; up to +28.29\% in BEV AP / +45.75\% in 3D AP compared to \textit{Direct Transfer}.

%% file: articles/conclusion.tex
\section{Conclusion}

\label{sec:conclusion}
In this work, we introduce a novel unsupervised domain adaptation approach, called {\framework}, to improve the generalization power of existing LiDAR-based 3D object detection models. 
To reduce the gap between source and target (where its labels are not accessible during training) domains, we propose two main steps: (i) Cross-modal LiDAR Encoder Pre-training and (ii) Cross-Domain LiDAR-Only Self-Training. In (i), a pair of image-based and LiDAR-based BEV features is aligned to learn modality-agnostic (and thus more domain-invariant) features. Further, in (ii), we apply an adversarial regularization to reduce the representation gap between source and target domains. Our extensive experiments on large-scale datasets demonstrate the effectiveness of our proposed method in various cross-domain adaptation scenarios, achieving state-of-the-art UDA performance.